\documentclass{article}

\usepackage[preprint]{nips_2018}

\usepackage[utf8]{inputenc} 
\usepackage[T1]{fontenc}    
\usepackage{hyperref}       
\usepackage{url}            
\usepackage{booktabs}       
\usepackage{amsfonts}       
\usepackage{nicefrac}       
\usepackage{microtype}      
\usepackage{graphicx}
\usepackage{subfigure}
\usepackage{amsmath}
\usepackage{algorithm}    
\usepackage{algorithmic}
\usepackage{array}
\usepackage{shortbold}

\title{Reinforced Continual Learning}

\author{
  Ju Xu  \\
  Center for Data Science, Peking University\\
  Beijing, China \\
  \texttt{xuju@pku.edu.cn} \\
  \And
   Zhanxing Zhu~\thanks{Corresponding author.}\\
   Center for Data Science, Peking University \& \\
   Beijing Institute of Big Data Research (BIBDR) \\
  Beijing, China \\
  \texttt{zhanxing.zhu@pku.edu.cn} \\
}

\begin{document}

\maketitle
\renewcommand\arraystretch{2}

\begin{abstract}
  Most artificial intelligence models   have  limiting ability to solve new tasks
  faster, without forgetting previously acquired knowledge. The recently emerging paradigm of continual learning aims to solve this issue, in which the model
  learns various tasks in a sequential fashion.
  In this work, a novel approach for continual learning is proposed,  which  searches for the best neural architecture for each coming task via sophisticatedly designed reinforcement learning strategies.  We name it as Reinforced Continual Learning. Our method not only has good performance on preventing catastrophic forgetting but also fits new tasks well.
  The experiments on sequential classification tasks for variants of MNIST and CIFAR-100 datasets demonstrate that the proposed approach outperforms existing continual learning alternatives for deep networks.
\end{abstract}

\section{Introduction}

Continual learning, or lifelong learning~\cite{thrun1}, the ability to learn
consecutive tasks without forgetting how to perform previously
trained tasks, is
an important topic for developing artificial intelligence. The primary goal of continual learning is to overcome the forgetting of learned tasks and to leverage  the earlier knowledge for obtaining better performance or faster convergence/training speed on the newly coming tasks.


In  deep learning community, two groups of strategies have been developed to alleviate the problem of  forgetting the previously trained tasks, distinguished by whether the network architecture changes during learning. 

The first category of approaches maintain a fixed network architecture with large capacity.  When training the network for consecutive tasks, some regularization term is enforced to prevent the model parameters from deviating too much
from the previous learned parameters according to their significance to old tasks~\cite{kirkpatrick1,zenke1}. In~\cite{lee2017overcoming}, the authors proposed to incrementally matches the moment of the posterior distribution of the neural network which is trained on the first and the second task, respectively. Alternatively, an episodic memory~\cite{GradientEpisodicMemory} is budgeted to store the subsets of previous datasets, and then trained together with the new task. 
Fernando et al. \cite{fernando1} proposed PathNet, in which a neural network has ten or twenty modules in each layer, and three or four modules are picked for one task in each layer by an evolutionary approach.  However, these methods typically require unnecessarily large-capacity networks, particularly when the number of tasks is  large, since the network architecture is never dynamically adjusted during training.

The other group of methods for overcoming catastrophic forgetting dynamically expand the network to accommodate the new coming task while keeping the parameters of previous architecture unchanging. Progressive networks~\cite{rusu1} expand the architectures with a fixed size of nodes or layers, leading to an extremely large network structure particularly faced with a large number of sequential tasks. The resultant complex architecture might be expensive to store and even unnecessary due to its high redundancy. Dynamically Expandable Network (DEN,~\cite{yoon1} alleviated this issue slightly by introducing group sparsity regularization when adding new parameters to the original network; unfortunately, there involves many hyperparameters in DEN, including various regularization and thresholding ones, which need to be tuned carefully due to the high sensitivity to the model performance.

In this work, in order to  better facilitate knowledge transfer and avoid catastrophic forgetting, we provide a novel framework to adaptively expand the network. Faced with a new task, deciding optimal number of nodes/filters to add for each layer is posed as a combinatorial optimization problem. We provide a sophisticatedly designed reinforcement learning method to solve this problem. Thus, we name it as Reinforced Continual Learning (RCL). In RCL, a controller implemented as a recurrent neural network is adopted to determine the best architectural hyper-parameters of neural networks for each task. We train the controller by an actor-critic strategy guided by a reward signal deriving from both validation accuracy and network complexity. This can maintain the prediction accuracy on older tasks as much as possible while reducing the overall model complexity.  To the best of our knowledge, the proposal is the first attempt that employs the reinforcement learning for solving the continual learning problems.

RCL not only differs from adding a fixed number of units to the old network for solving a new task~\cite{rusu1}, which might be suboptimal and computationally expensive, but also distinguishes from~\cite{yoon1} as well that performs group sparsity regularization on the added  parameters.  We validate the effectiveness of RCL on various sequential tasks. And the results show that RCL can obtain better performance than existing methods even with adding much less units.

The rest of this paper is organized as follows. In Section~\ref{sec:pre}, we introduce the preliminary knowledge on reinforcement learning. We propose the new method RCL in Section~\ref{sec:rcl}, a model to learn a sequence of tasks dynamically based on reinforcement learning. In Section~\ref{sec:exp}, we implement various experiments to demonstrate the superiority of  RCL over other state-of-the-art methods. Finally, we conclude our paper in Section~\ref{sec:con} and provide some directions for future research.

\section{Preliminaries of Reinforcement learning} 
\label{sec:pre}
Reinforcement learning \cite{sutton1} deals with learning a policy for an
agent interacting in an unknown environment. It has been applied successfully to various problems, such as games \cite{minh1,silver1}, natural language processing \cite{yu1}, neural architecture/optimizer search \cite{zoph1,bello1} and so on. At each step,
an agent observes the current state $s_t$ of the environment,
decides of an action $a_t$ according to a policy $\pi(a_t | s_t)$, and observes
a reward signal $r_{t+1}$. The goal of the agent is to find a policy
that maximizes the expected sum of discounted rewards $R_t$,
$R_t = \sum_{t'=t+1}^\infty \gamma^{t'-t-1 }r_{t'},$
where $\gamma \in (0,1]$ is a discount factor that determines the importance of
future rewards. 
 The value function of a policy $\pi$ is defined as the expected return $V_{\pi}(s) = E_\pi[\sum_{t=0}^\infty\gamma^{t}r_{t+1}|s_0 = s]$ and its action-value function as $Q_{\pi}(s,a) = E_\pi[\sum_{t=0}^\infty\gamma^{t}r_{t+1}|s_0 = s, a_0 =a]$.

Policy gradient methods address the problem of finding a good policy by performing stochastic gradient descent to optimize a performance objective over a given family of parametrized stochastic policies $\pi_\theta (a|s)$ parameterized by $\theta$. The policy
gradient theorem \cite{sutton2} provides expressions for the gradient of the average reward and discounted reward objectives with respect to $\theta$. In the discounted setting, the objective is defined with respect to a designated start state (or distribution) $s_0$: $\rho(\theta,s_0) = E_{\pi_\theta}[\sum_{t=0}^\infty\gamma^{t}r_{t+1}|s_0 ]$. The policy gradient theorem shows
that: 
\begin{alignat}{2}  
\label{eq01}
\frac{\partial \rho(\theta,s_0)}{\partial \theta} = \sum_s \mu_{\pi_\theta}(s|s_0)\sum_a\frac{\partial \pi_{\pi_\theta}(a|s)}{\partial \theta}Q_{\pi_\theta}(s,a).
\end{alignat}
where $\mu_{\pi_\theta}(s|s_0)= \sum_{t=0}^\infty\gamma^tP(s_t = s|s_0) $.

\section{Our Proposal: Reinforced Continual Learning}
\label{sec:rcl}
In this section, we elaborate on the new framework for continual learning, Reinforced Continual Learning(RCL). RCL consists of three networks,  \emph{controller},  \emph{value network}, and  \emph{ task network}.  The controller is implemented as a Long Short-Term Memory network (LSTM) for generating policies and determining how many filters or nodes will be added for each task. We design the value network as a fully-connected network, which approximates the value of the state. The task network can be any network of interest for solving a particular task, such as image classification or object detection.  In this paper, we  use a convolutional network (CNN) as the task network to demonstrate how RCL adaptively expands this CNN to prevent forgetting, though our method can not only adapt to convolutional networks, but also to fully-connected networks.

\subsection{The Controller}
\begin{figure}{ht}
		\centering
    \begin{tabular}{cc}
		\includegraphics[width=2.2in]{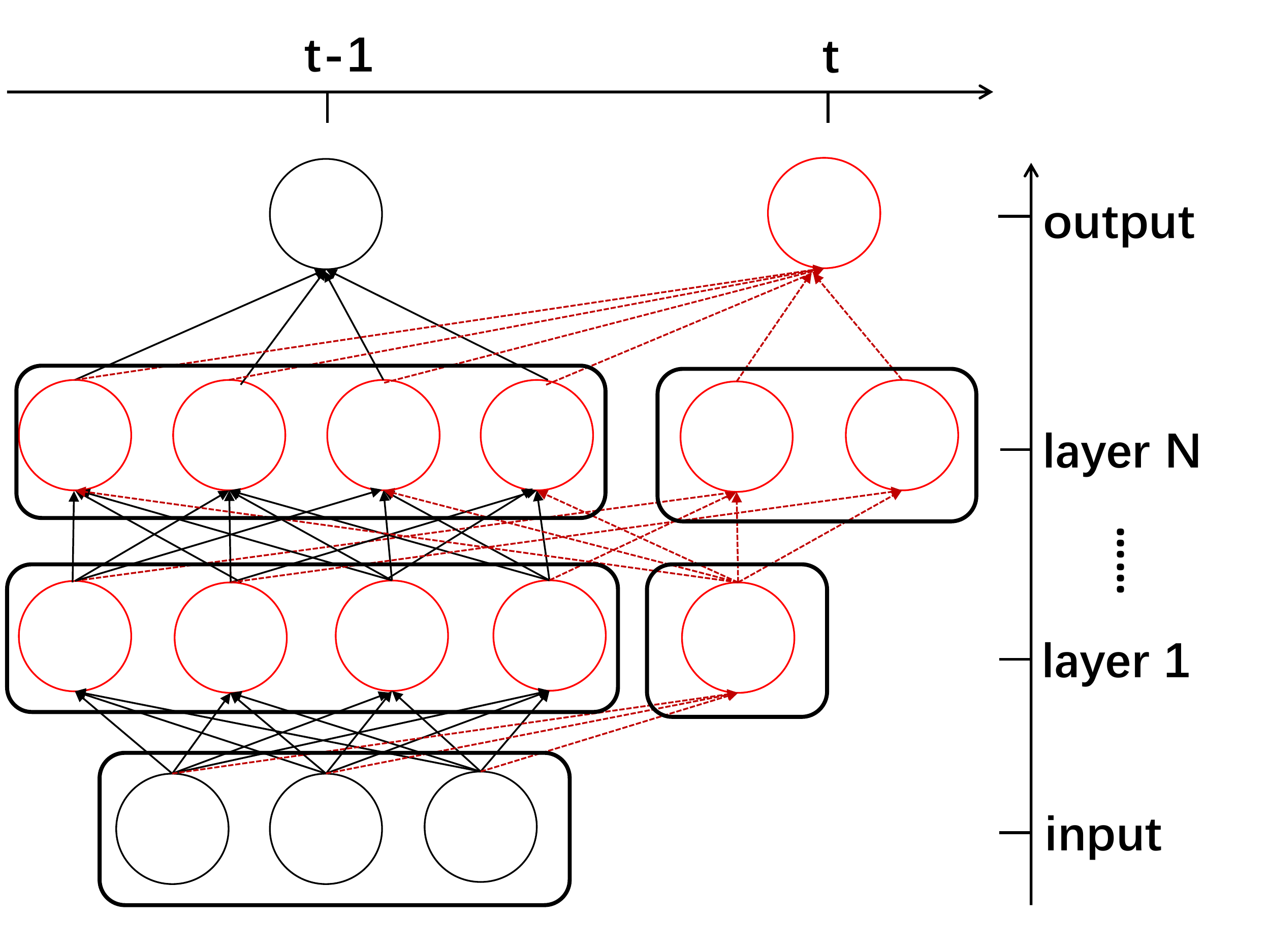} & \includegraphics[width=2.2in]{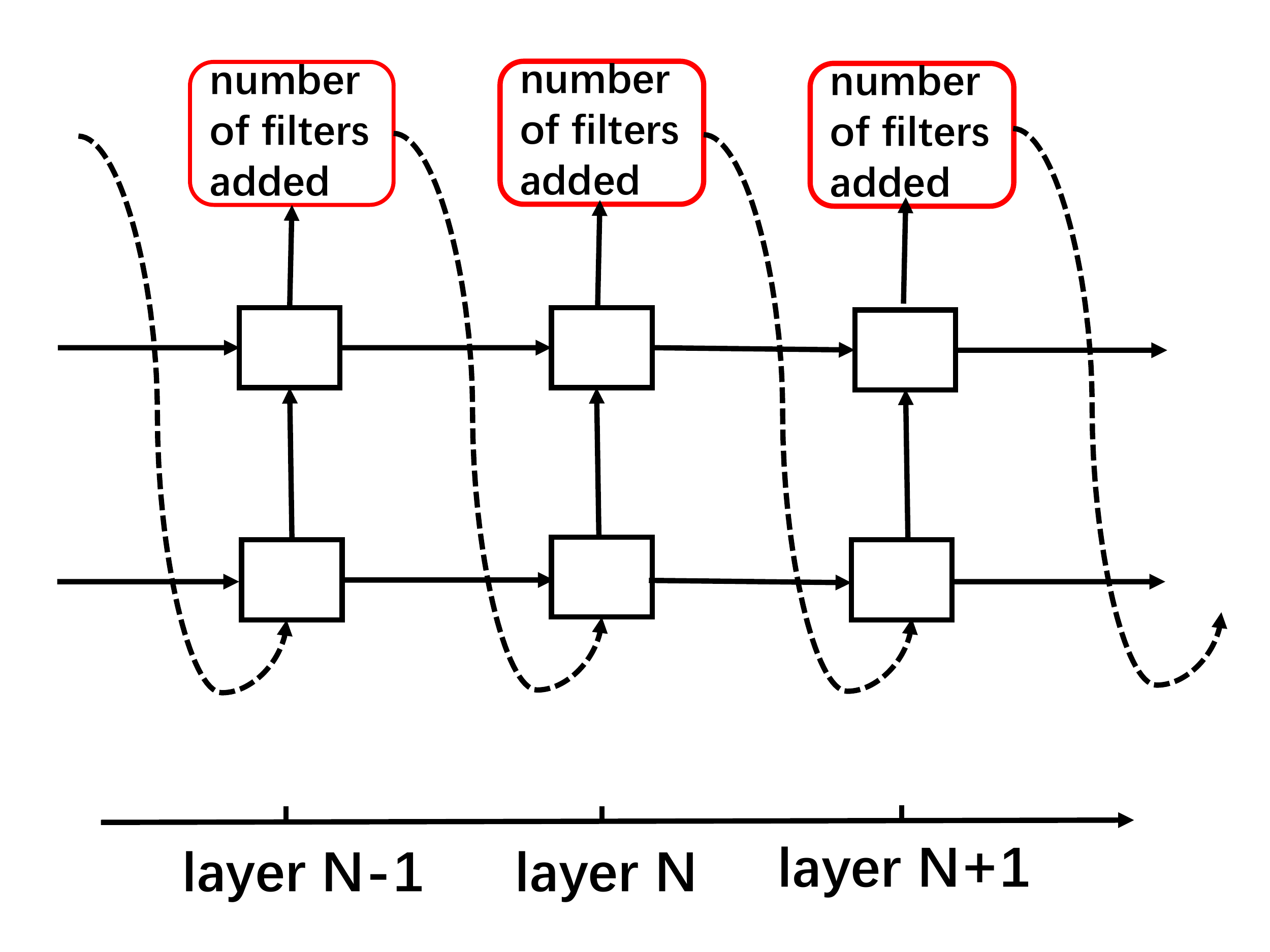} \\ 
    (a) & (b)
    \end{tabular}
    \vspace{-0.3cm}
 		 \caption{(a) RCL adaptively expands the each layer of network when $t$-th task arrives. (b) The controller implemented as a RNN to determine how many filters to add for the new task.}
		\label{fig:rcl}
\end{figure}

Figure ~\ref{fig:rcl}(a) visually shows how RCL expands the network when a new task arrives. After the learning process of task $t-1$ finishes and task $t$ arrives, we use a controller to decide how many filters or nodes should be added to each layer. In order to prevent semantic drift, we withhold modification of network weights for previous tasks and only train the newly added filters. After we have trained the model for task $t$, we timestamp each newly added filter by the shape of every layer to prevent the caused semantic drift. During the inference time, each task only employs the parameters introduced in stage $t$,  and does not consider the new filters added in the later tasks.

Suppose the task network has $m$ layers, when faced with a newly coming task, for each layer $i$, we specify the the number of filters to add in the range between $0$ and $n_i-1$. A straightforward idea to obtain the optimal configuration of added filters for  $m$ layers  is to traverse all the  combinatorial combinations of actions. However, for an $m$-layer network, the time complexity of collecting the best action combination is $\mathcal{O}(\prod_1^m n_i)$, which is NP-hard and unacceptable for very deep architectures such as VGG and ResNet. 

To deal with this issue, we treat a series of actions as a fixed-length string. It is possible to use a controller to generate such a string, representing how many filters should be added in each layer. 
Since there is a recurrent relationship between consecutive layers, the controller can be naturally  designed  as a LSTM network.  At the first step, the controller
network receives an empty embedding as input (i.e. the state $s$) for the current task, which will be fixed during the training.  
For each task $t$, we equip the network with softmax output, $\pB_{t,i} \in \Rbb^{n_i}$ representing the probabilities of sampling each action for layer $i$, i.e. the number of filters to be added.  We design the LSTM in an autoregressive manner,  as Figure~\ref{fig:rcl}(b) shows, the probability $p_{t,i}$ in the previous step is fed as input into the next step. This process is circulated until we obtain  the actions and probabilities for  all the $m$ layers. And the policy probability of the sequence of actions $a_{1:m}$ follows the product rule,  
\begin{equation}
\pi(a_{1:m}|s;\theta_c) = \prod_{i=1}^{m} p_{t,i,a_i},
\end{equation}
 where $\theta_c$ denotes the parameters of the controller network. 

\subsection{The Task Network}
We deal with  $T$ tasks arriving in a sequential manner  with training dataset $\Dcal_t=\{x_i,y_i\}_{i=1}^{N_t}$ , validation dataset $\Vcal_t=\{x_i,y_i\}_{i=1}^{M_t}$, test dataset $\Tcal_t = \{x_i,y_i\}_{i=1}^{K_t}$ at time $t$.  
For the first task, we train a basic task network that performs well enough via solving a standard supervised learning problem,
\begin{alignat}{2}
\label{init}   
\min_{W_1} L_1(W_1;\Dcal_1).
\end{alignat}
We define the well-trained parameters as $W_t^a$ for task $t$.  
When the $t$-th task arrives, we have already known the best parameters $W_{t-1}^a$ for task $t-1$. Now we use the controller to decide how many filters should be added to each layer, and then we obtain an expanded child network, whose parameters to be learned are denoted as $W_t$ (including $W_{t-1}^a$). The training procedure for the new task is as follows, keeping $W_{t-1}^a$ fixed and only back-propagating the newly added parameters of $W_t \backslash W_{t-1}^a$. Thus, the optimization formula for the new task is,
\begin{alignat}{2}
\label{latter}   
\min_{W_t \backslash W_{t-1}^a} L_t(W_t;\Dcal_t).
\end{alignat}
We use stochastic gradient descent to learn the newly added filters with $\eta$ as the learning rate,
\begin{alignat}{2}
\label{update}
& W_t \backslash W_{t-1}^a \longleftarrow W_t \backslash W_{t-1}^a - \eta \nabla_{W_t \backslash W_{t-1}^a} L_t.
\end{alignat}
The expanded child network will be trained until  the required number of epochs or convergence are reached. And then we test the child network on the validation dataset $\Vcal_t$ and the corresponding accuracy $A_t$ will be returned. 
The parameters of the expanded network achieving the maximal reward (described in Section~\ref{sec:reward}) will be the optimal ones for task $t$, and we store them for later tasks.

\subsection{Reward Design}
\label{sec:reward}
In order to facilitate  our controller to generate better actions over time, we need design a reward function to reflect the performance of our actions.  Considering both the validation accuracy and complexity of the expanded network, we design the reward for task $t$ by the combination of the two terms, 
\begin{equation}
R_t = A_t(S_t,a_{1:m})  + \alpha C_t(S_t,a_{1:m}), \label{eq:reward}
\end{equation}
where $A_t$ represents the validation accuracy on $\Vcal_t$, the network complexity as $C_t = \sum\limits_{i=1}^m k_i$, $k_i$ is the numbers of filters added in layer $i$, and $\alpha$ is a parameter to balance between the prediction performance and model complexity.  
Since $R_t$ is non-differentiable, we use policy gradient to update the controller, described in the following section.

\subsection{Training Procedures} 

The controller's prediction can be viewed as a list of actions $a_{1:m}$, which means the number of filters added in $m$ layers , to design an
new architecture for a child network and then be trained in a new task. At convergence, this child network will achieve an accuracy $A_t$ on
a validation dataset and the model complexity $C_t$, finally we can obtain the reward $R_t$ as defined in Eq.~\eqref{eq:reward}. We can use this reward $R_t$ and reinforcement learning
to train the controller. 

To find the optimal incremental architecture the new task $t$, the controller aims  to maximize its expected reward,
\begin{alignat}{2}
J(\theta_c)=V_{\theta_c}(s_t).
\end{alignat}
where $V_{\theta_c}$ is the true value function. 
 In order to accelerate policy gradient training over $\theta_c$, we use actor–critic methods with
a value network parameterized by $\theta_v$ to approximate the state value $V(s_t; \theta_v)$.   The REINFORCE algorithm~\cite{william1} can be used to learn $\theta_c$,
\begin{align}
\label{reinforce1}
\nabla_{\theta_c}J(\theta_c) = E \left[ \sum \limits_{a_{1:m}} \pi(a_{1:m}|s_t,\theta_c)(R(s_t,a_{1:m})-V(s_t,\theta_v))\frac{\nabla_{\theta_c}\pi(a_{1:m}|s_t,\theta_c)}{\pi(a_{1:m}|s_t,\theta_c)} \right].
\end{align}

A Monte Carlo approximation for the above quantity is,
\begin{alignat}{2}
\label{policy}
\frac{1}{N}\sum_{i=1}^N\nabla_{\theta_c}\log \pi(a_{1:m}^{(i)}|s_t;\theta_c)\left(R(s_t,a_{1:m}^{(i)})-V(s_t,\theta_v)\right).
\end{alignat}
where $N$ is the batch size. For the value network, we utilize  gradient-based method to update $\theta_v$, the gradient of which can be evaluated as follows, 
\begin{align}
\label{valuenet}
&L_v = \frac{1}{N}\sum_{i=1}^N(V(s_t;\theta_v) - R(s_t,a^{(i)}_{1:m}))^2, \nonumber \\
&\nabla_{\theta_v} L_v = \frac{2}{N}\sum_{i=1}^N\left( V(s_t;\theta_v) - R(s_t,a^{(i)}_{1:m}) \right) \frac{\partial V(s_t;\theta_v)}{\partial \theta_v}.
\end{align}

\begin{algorithm}[tb]
  \caption{RCL for Continual Learning}
  \label{alg:rcl}
  \begin{algorithmic}[1]
    \STATE {\bfseries Input:} A sequence of dataset $\Dcal=\{ \Dcal_1, \Dcal_2,\dots,\Dcal_T \}$ 
    \STATE {\bfseries Output:} $W_T^a$
    \FOR{$t=1,...,T $}
    \IF {$t=1$}
    \STATE Train the base network using (~\ref{init}) on the first datasest $\Dcal_1$ and obtain $W_1^a$.
    \ELSE
    \STATE Expand  the network by Algorithm~\ref{alg:expanding}, and obtain the trained  $W_t^a$.
    \ENDIF
    \ENDFOR
  \end{algorithmic}
\end{algorithm}

\begin{algorithm}[tb]
  \caption{Routine for Network Expansion}
  \label{alg:expanding}
  \begin{algorithmic}[1]
    \STATE {\bfseries Input:} Current dataset $\Dcal_t$; previous parameter $W_{t-1}^a$; the size of action space for each layer $n_i, i=1\dots,m$; number of epochs for training the controller and value network, $T_e$.
    \STATE {\bfseries Output:} Network parameter $W_t^a$
    \FOR{$i=1,\dots,T_e $}
    \STATE Generate actions $a_{1:m}$ by controller's policy;
    \STATE Generate $W_t^{(i)}$ by expanding parameters $W_{t-1}^a$ according to $a_{1:m}$; 
    \STATE Train the expanded network using Eq.~\eqref{update} to obtain $W_t^{(i)}$.
    \STATE Evaluate the gradients of the controller and value network by Eq.~\eqref{policy} and Eq.\eqref{valuenet},
    \begin{align}
    \theta_c = \theta_c + \eta_c \nabla_{\theta_c}J(\theta_c), \quad
    \theta_v = \theta_v - \eta_v \nabla_{\theta_v}L_v(\theta_v). \nonumber
    \end{align}
    \ENDFOR
    \STATE Return the best network parameter configuration, $W_t^a = \argmax_{W_t^{(i)}} R_t(W_t^{(i)})$.
  \end{algorithmic}
\end{algorithm}

Finally we summarize our RCL approach for continual learning in Algorithm~\ref{alg:rcl}, in which the subroutine for network expansion is described in Algorithm~\ref{alg:expanding}. 

\subsection{Comparison with Other Approaches}

As a new framework for network expansion to achieve continual learning, RCL distinguishes from progressive network~\cite{rusu1} and DEN~\cite{yoon1} from the following aspects. 
\begin{itemize}
\item Compared with DEN, instead of performing selective retraining and network split,  RCL keeps the learned parameters for previous tasks fixed and only updates the added parameters. Through this training strategy, RCL can totally prevent catastrophic forgetting due to the freezing parameters for corresponding tasks. 
\item Progressive neural networks expand the architecture with a fixed number of units or filters. To obtain a satisfying model accuracy when number of sequential tasks is large, the final complexity of progressive nets is required to be extremely high. This directly leads to high computational burden both in training and inference, even difficult for the storage of the entire model. To handle this issue, both RCL and DEN dynamically adjust the networks to reach a more economic architecture.  
\item While DEN achieves the expandable network by sparse regularization, RCL adaptively expands the network by reinforcement learning. 
However, the performance of DEN is quite sensitive to the various hyperparameters, including regularization parameters and thresholding coefficients. RCL largely reduces the number of hyperparameters and boils down to only balancing the average validation accuracy and model complexity when the designed reward function.  Through different experiments in Section~\ref{sec:exp}, we demonstrate that RCL could achieve more stable results, and better model performance could be achieved simultaneously with even much less neurons than DEN.
\end{itemize}

\section{Experiments}
\label{sec:exp}
We perform a variety of experiments to access the performance of RCL in continual learning. We will report the accuracy, the model complexity and the training time consumption between our RCL and the state-of-the-art baselines. We implemented all the experiments in Tensorfolw framework on GPU Tesla K80. 

\paragraph{Datasets}
(1) MNIST Permutations~\cite{kirkpatrick1}. Ten variants of the MNIST data, where each task is transformed by a fixed permutation of
pixels. In the dataset, the samples from different task are not independent and identically distributed;
(2) MNIST Mix. Five MNIST permutations ($P_1,\dots,P_5$) and five variants of the MNIST dataset ($R_1,\dots,R_5$) where each contains digits rotated by a fixed angle between 0 and 180 degrees. These tasks are arranged in the order $P_1,R_1, P_2,\dots,P_5,R_5$.  
(3) Incremental CIFAR-100~\cite{icart}. Different from the original CIFAR-100, each task introduces a new set of classes. For the total number of tasks $T$, each new task contains digits from a subset of $100/T$ classes. In this dataset, the distribution of the input is similar for all tasks, but the distribution of the output is different.

For all of the above datasets, we set the number of tasks to be learned as $T = 10$. For the MNIST datasets, each task contains 60000 training examples and 1000 test examples from 10 different classes. For the CIFAR-100 datasets, each task contains 5000 train examples and 1000 examples from 10 different classes. The model observes the tasks one by one, and once the task had been observed, the task will not be observed later during the training.

\paragraph{Baselines}
(1) \textbf{SN}, a single network trained across all tasks; 
(2) \textbf{EWC}, deep network trained with elastic weight consolidation~\cite{kirkpatrick1} for regularization; 
(3) \textbf{GEM}, gradient episodic memory~\cite{GradientEpisodicMemory};
(4) \textbf{PGN}, progressive neural network proposed in~\cite{rusu1};
(5) \textbf{DEN}, dynamically expandable network~\cite{yoon1}.

\paragraph{Base network settings}
(1) Fully con{}nected networks for  MNIST Permutations and MNIST Mix datasets. We use a three-layer network with 784-312-128-10 neurons with RELU activations;
(2) LeNet is used for Incremental CIFAR-100. LeNet has two convolutional layers and three fully-connected layers, the detailed structure of {}LeNet can be found in~\cite{lenet1}.

\begin{figure}[htbp]
  \centering
    \begin{minipage}[b]{0.85\textwidth}
      \includegraphics[width=1\textwidth]{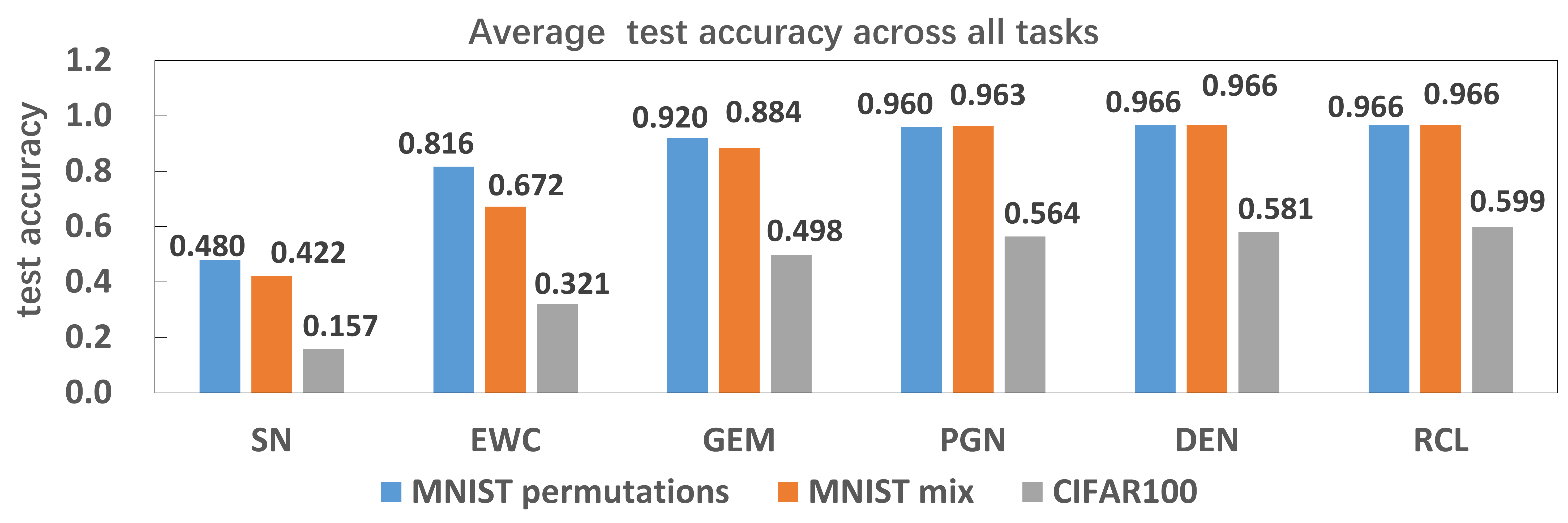}
    \end{minipage}
      \begin{minipage}[b]{0.85\textwidth}
    \includegraphics[width=1\textwidth]{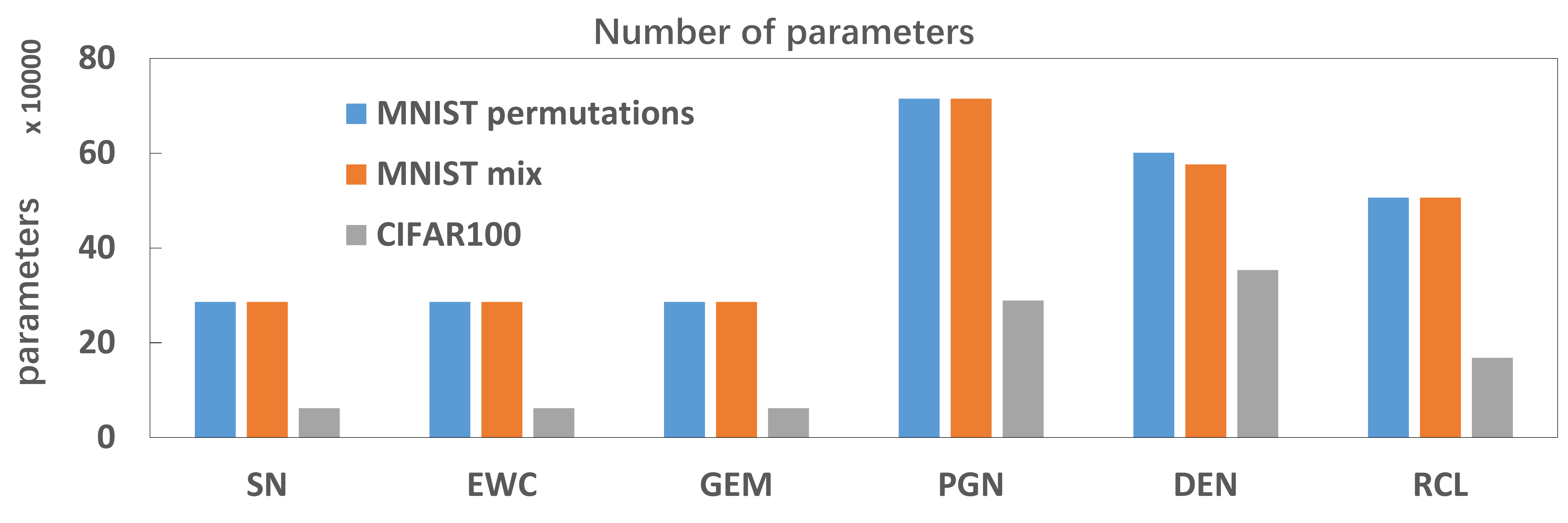}
  \end{minipage}
  \caption{Top: Average test accuracy for all the datasets. Bottom: The number of parameters  for different methods. } \label{experiments}
\end{figure}

\begin{figure}[htbp]
  \includegraphics[width=1\textwidth]{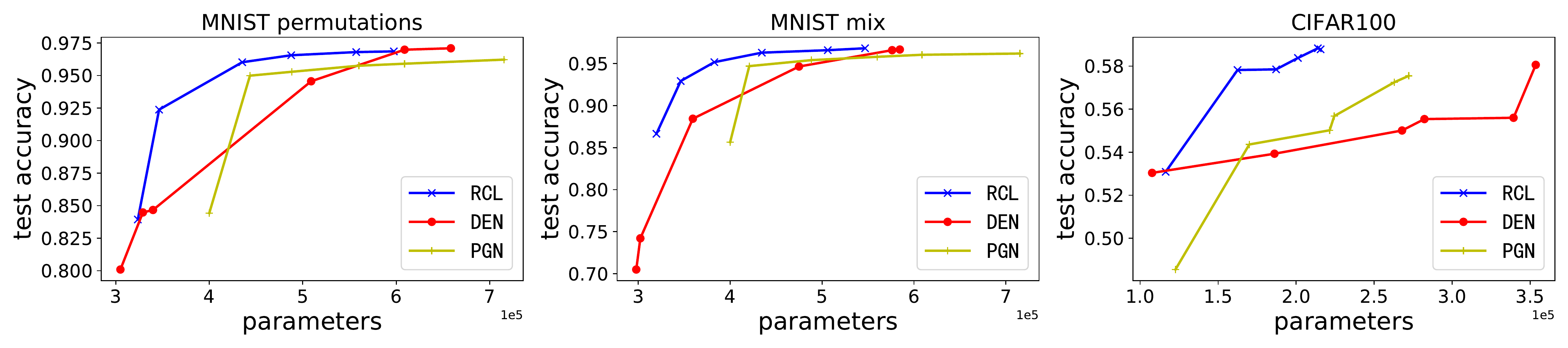}
  \vspace{-0.5cm}
  \caption{Average test accuracy \emph{v.s.} model complexity for RCL, DEN and PGN.}
  \label{fig:accvspara}
\end{figure}

\begin{figure}[htbp]
  \centering
      \includegraphics[width=1\textwidth]{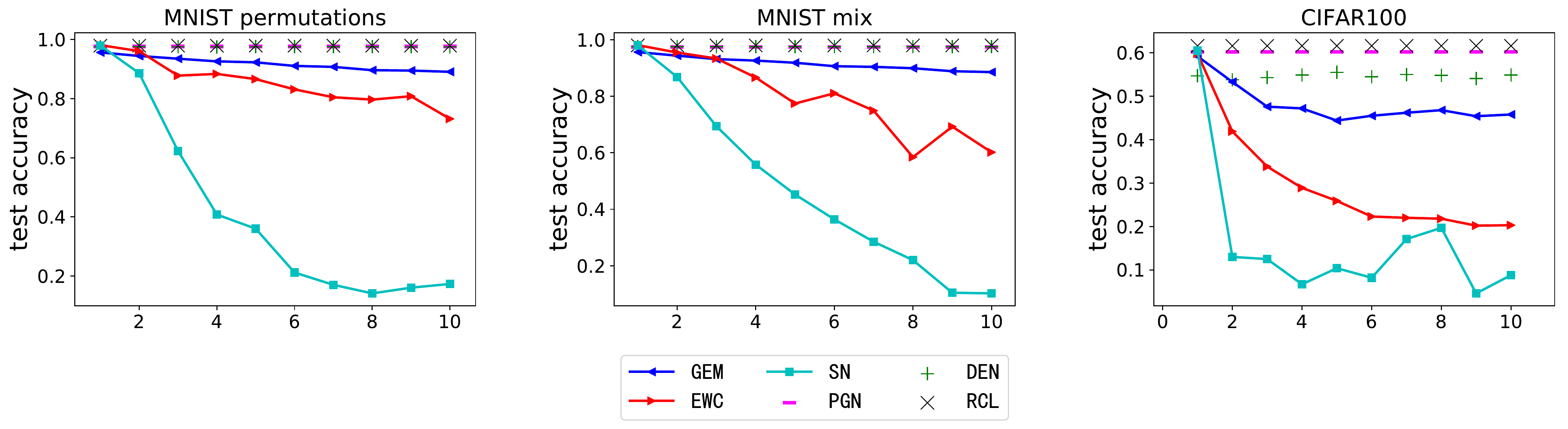}
      \vspace{-0.5cm}
  \caption{Test accuracy on the first task as more tasks are learned.} \label{fig:forgetting}
\end{figure}

\subsection{Results}
We evaluate each compared approach by considering  average test accuracy on all the tasks, model complexity and training time. Model complexity is measured via the number of model parameters after training all the tasks. We first report the test accuracy and model complexity of baselines and our proposed RCL for the three datasets in Figure~\ref{experiments}. 

\paragraph{Comparison between fixed-size and expandable networks.}
From Figure~\ref{experiments}, we can easily observe that the approaches with fixed-size network architectures, such as  IN, EWC and GEM, own low model complexity, but their prediction accuracy is much worse than those methods with expandable networks, including PGN, DEN and RCL. This shows that dynamically expanding networks can  indeed contribute to the model performance by a large margin. 

\paragraph{Comparison between PGN, DEN and RCL.} Regarding to the expandable networks, RCL outperforms PGN and DEN on on both test accuracy and model complexity. Particularly, RCL achieves significant reduction on the number of parameters compared with PGN and DEN, e.g. for incremental Cifar100 data, $42\%$ and $53\%$ parameter reduction, respectively. 

To further see the difference of the three methods, we vary the hyperparameters settings and train the networks accordingly, and obtain how test accuracy changes with respect to the number of parameters, as shown in Figure~\ref{fig:accvspara}. We can clearly observe that RCL can achieve significant model reduction with the same test accuracy as that of PGN and DEN, and remarkable accuracy improvement with same size of networks.   
This demonstrates the benefits of employing reinforcement learning to adaptively control the complexity of the entire model architecture.

\paragraph{Evaluating the forgetting behavior.} 
Figure~\ref{fig:forgetting} shows the evolution of the test accuracy on the first task as more tasks are learned. RCL and PGN exhibit no forgetting while the approaches without expanding the networks raise catastrophic forgetting. Moreover, DEN can not completely prevent forgetting since it retrains the previous parameters when learning new tasks.

\paragraph{Training time}
We report the wall clock training time for each compared method in Table~\ref{table1}). Since RCL is based on reinforcement learning, a large number of trials are typically required that leads to more training time than other methods. Improving the training efficiency of reinforcement learning is still an open problem, and we leave it as future work.


\begin{table}[ht]
  \caption{Training time (in seconds) of experiments for all methods.}
  \centering
  \begin{tabular}{lcccccccc}
    \toprule
    Methods &IN&EWC&GEM&DEN&PGN&RCL \\
    \hline
    MNIST permutations&173&1319&1628&21686&452&34583\\
    \hline
    MNIST mix&170&1342&1661&19690&451&23626\\
    \hline
    CIFAR100&149&508&7550&1428&167&3936\\
    \bottomrule
  \end{tabular}
  \label{table1}
\end{table}

\begin{figure}[htbp]
  \includegraphics[width=1\textwidth]{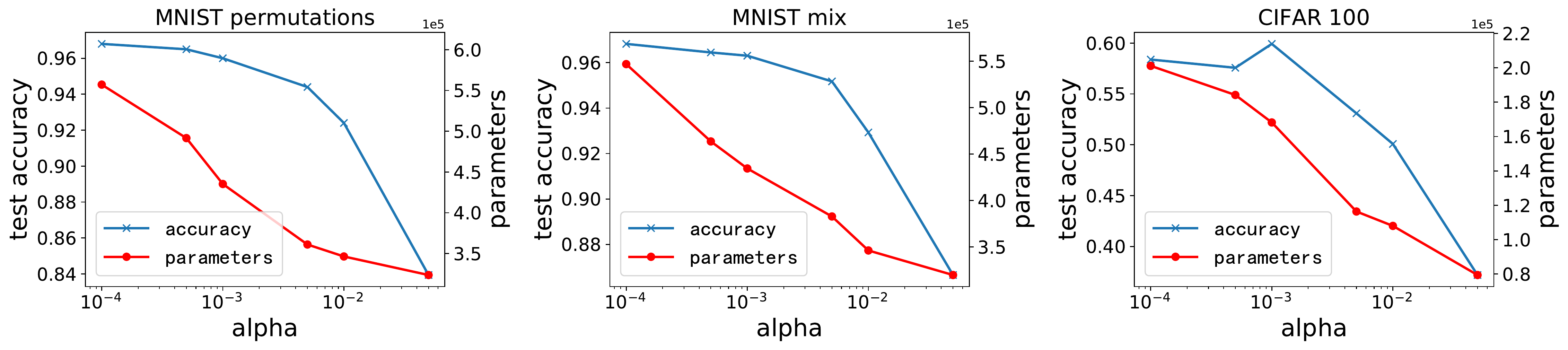}
  \vspace{-0.5cm}
  \caption{Experiments on the influence of the parameter $\alpha$ in the reward design}
  \label{fig:alpha}
\end{figure}

\paragraph{Balance between test accuracy and model complexity.}
We control the tradeoff between the model performance and complexity through the coefficient $\alpha$ in the reward function~\eqref{eq:reward}. Figure~\ref{fig:alpha} shows how varying $\alpha$ affects the test accuracy and number of model parameters. As expected, with increasing $\alpha$ the model complexity drops significantly while the model performance also deteriorate gradually.  Interestingly, when $\alpha$ is small, accuracy drops much slower compared with the decreasing of the number of parameters. This observation could help to choose a suitable $\alpha$ such that a medium-sized network can still achieve a relatively good model performance. 

\section{Conclusion}
\label{sec:con}
We have proposed a novel framework for continual learning, Reinforced Continual Learning. Our method searches for the best neural architecture for coming task by reinforcement learning, which increases its capacity when necessary and effectively prevents semantic drift. We implement both fully connected and convolutional neural networks as our task networks, and validate them on different datasets. The experiments demonstrate that our proposal outperforms the exiting baselines significantly both on prediction accuracy and model complexity.

As for future works,  two directions are worthy of consideration. Firstly, we will develop new strategies for RCL to facilitate  backward transfer, i.e. improve previous tasks' performance by learning new tasks. Moreover, how to reduce the training time of RCL is particularly important for large networks with more layers.


\bibliography{rcl}

\begin{thebibliography}{10}

\bibitem{bello1}
Irwan Bello, Barret Zoph, Vijay Vasudevan, and Quoc~V. Le.
\newblock Neural optimizer search with reinforcement learning.
\newblock In {\em International Conference on Machine Learning(ICML)}, pages
  459--468, 2017.

\bibitem{fernando1}
Chrisantha Fernando, Dylan Banarse, Charles Blundell, Yori Zwols, David Ha,
  Andrei~A. Rusu, Alexander Pritzel, and Daan Wierstra.
\newblock Pathnet: Evolution channels gradient descent in super neural
  networks.
\newblock {\em arXiv preprint arXiv:1701.08734}, 2017.

\bibitem{goodfe1}
Ian~J Goodfellow, Mehdi Mirza, Da~Xiao, Aaron Courville, and Yoshua Bengio.
\newblock An empirical investigation of catastrophic forgetting in
  gradient-based neural networks.
\newblock {\em arXiv preprint arXiv:1312.6211}, 2013.

\bibitem{kirkpatrick1}
James Kirkpatrick, Razvan Pascanu, Neil~C. Rabinowitz, Joel Veness, Guillaume
  Desjardins, Andrei~A. Rusu, Kieran Milan, John Quan, Tiago Ramalho, Agnieszka
  Grabska{-}Barwinska, Demis Hassabis, Claudia Clopath, Dharshan Kumaran, and
  Raia Hadsell.
\newblock Overcoming catastrophic forgetting in neural networks.
\newblock {\em Proceedings of the National Academy of Sciences},
  114(13):3521--3526, 2017.

\bibitem{Krizhevsky2009LearningML}
Alex Krizhevsky.
\newblock Learning multiple layers of features from tiny images.
\newblock 2009.

\bibitem{kumar1}
Abhishek Kumar and Hal {Daum\'{e} III}.
\newblock Learning task grouping and overlap in multi-task learning.
\newblock In {\em International Conference on Machine Learning(ICML),
  Edinburgh, Scotland, UK}, 2012.

\bibitem{lenet1}
Yann LeCun, L\'{e}on Bottou, Yoshua Bengio, and Patrick Haffner.
\newblock Gradient-based learning applied to document recognition.
\newblock {\em Proceedings of the IEEE}, 86(11):2278--2324, 1998.

\bibitem{lee1}
Sang{-}Woo Lee, Jin{-}Hwa Kim, Jaehyun Jun, Jung{-}Woo Ha, and Byoung{-}Tak
  Zhang.
\newblock Overcoming catastrophic forgetting by incremental moment matching.
\newblock In {\em In Advances in Neural Information Processing Systems}, pages
  4655--4665, 2017.

\bibitem{lee2}
Sang{-}Woo Lee, Chung{-}yeon Lee, Dong{-}Hyun Kwak, Jiwon Kim, Jeonghee Kim,
  and Byoung{-}Tak Zhang.
\newblock Dual-memory deep learning architectures for lifelong learning of
  everyday human behaviors.
\newblock In {\em International Joint Conference on Artificial Intelligence
  (IJCAI)}, pages 1669--1675, 2016.

\bibitem{GradientEpisodicMemory}
David Lopez-Paz and Marc'Aurelio Ranzato.
\newblock Gradient episodic memory for continual learning.
\newblock In {\em NIPS}, 2017.

\bibitem{minh1}
Volodymyr Mnih, Koray Kavukcuoglu, David Silver, Andrei~A. Rusu, Joel Veness,
  Marc~G. Bellemare, Alex Graves, Martin~A. Riedmiller, Andreas Fidjeland,
  Georg Ostrovski, Stig Petersen, Charles Beattie, Amir Sadik, Ioannis
  Antonoglou, Helen King, Dharshan Kumaran, Daan Wierstra, Shane Legg, and
  Demis Hassabis.
\newblock Human-level control through deep reinforcement learning.
\newblock {\em Nature}, 518(7540):529--533, 2015.

\bibitem{icart}
Sylvestre{-}Alvise Rebuffi, Alexander Kolesnikov, Georg Sperl, and Christoph~H.
  Lampert.
\newblock icarl: Incremental classifier and representation learning.
\newblock In {\em {CVPR}}, pages 5533--5542. {IEEE} Computer Society, 2017.

\bibitem{rusu1}
Andrei~A. Rusu, Neil~C. Rabinowitz, Guillaume Desjardins, Hubert Soyer, James
  Kirkpatrick, Koray Kavukcuoglu, Razvan Pascanu, and Raia Hadsell.
\newblock Progressive neural networks.
\newblock {\em arXiv preprint arXiv:1606.04671}, 2016.

\bibitem{progressive}
Andrei~A. Rusu, Neil~C. Rabinowitz, Guillaume Desjardins, Hubert Soyer, James
  Kirkpatrick, Koray Kavukcuoglu, Razvan Pascanu, and Raia Hadsell.
\newblock Progressive neural networks.
\newblock {\em CoRR}, abs/1606.04671, 2016.

\bibitem{ruvolo1}
Paul Ruvolo and Eric Eaton.
\newblock {ELLA:} an efficient lifelong learning algorithm.
\newblock In {\em International Conference on Machine Learning(ICML), Atlanta,
  GA, USA}, pages 507--515, 2013.

\bibitem{silver1}
David Silver, Julian Schrittwieser, Karen Simonyan, Ioannis Antonoglou, Aja
  Huang, Arthur Guez, Thomas Hubert, Lucas Baker, Matthew Lai, Adrian Bolton,
  et~al.
\newblock Mastering the game of go without human knowledge.
\newblock {\em Nature}, 550(7676):354, 2017.

\bibitem{sriva1}
Rupesh~Kumar Srivastava, Jonathan Masci, Sohrob Kazerounian, Faustino~J. Gomez,
  and J{\"{u}}rgen Schmidhuber.
\newblock Compete to compute.
\newblock In {\em Advances in Neural Information Processing Systems}, pages
  2310--2318, 2013.

\bibitem{sutton1}
Richard~S. Sutton and Andrew~G. Barto.
\newblock {\em Reinforcement Learning: An Introduction}.
\newblock Cambridge: MIT press, 1998.

\bibitem{sutton2}
Richard~S. Sutton, David~A. McAllester, Satinder~P. Singh, and Yishay Mansour.
\newblock Policy gradient methods for reinforcement learning with function
  approximation.
\newblock In {\em Advances in Neural Information Processing Systems}, pages
  1057--1063, 1999.

\bibitem{thrun1}
Sebastian Thrun.
\newblock A lifelong learning perspective for mobile robot control.
\newblock In {\em International Conference on Intelligent Robots and Systems},
  1995.

\bibitem{william1}
Ronald~J. Williams.
\newblock Simple statistical gradient-following algorithms for connectionist
  reinforcement learning.
\newblock {\em Machine Learning}, 8:229--256, 1992.

\bibitem{yoon1}
J.~Yoon and E.~Yang.
\newblock Lifelong learning with dynamically expandable networks.
\newblock {\em arXiv preprint arXiv:1708.01547}, 2017.

\bibitem{yu1}
Lantao Yu, Weinan Zhang, Jun Wang, and Yong Yu.
\newblock Seqgan: Sequence generative adversarial nets with policy gradient.
\newblock In {\em {AAAI}}, pages 2852--2858, 2017.

\bibitem{zenke1}
Friedemann Zenke, Ben Poole, and Surya Ganguli.
\newblock Continual learning through synaptic intelligence.
\newblock In {\em International Conference on Machine Learning (ICML)}, 2017.

\bibitem{zoph1}
Barret Zoph and Quoc~V. Le.
\newblock Neural architecture search with reinforcement learning.
\newblock {\em arXiv preprint arXiv:1611.01578}, 2016.

\end{thebibliography}
\bibliographystyle{plain}

\appendix

\section{Experiment settings}
In this section, we will present the experiments details of our model and baselines. When dealing with dataset MNIST permutations and dataset MNIST mix, we use a three-layer network with 784-312-128-10 neurons, and the learning rate is 0.001, the batch size is 32, the training epochs are 15 for all models. When expanding the network, the size of search space is 30 across all layers for RCL,DEN and PGN. As for CIFAR-100, we use LeNet as our task network. The training epochs are 20 and the learning rate is 0.001. The search space is 5 in convolutional layers, 25 in fully-connected layers for RCL,DEN and PGN. \\
\\
Our controller is implemented as a LSTM network. The LSTM network has two layers, and the hidden size is 100. Our value network is implemented as a fully-connected network, which has only one layer. The learning rate for our controller is 0.001, for our value network is 0.005. \\
\\
The $\alpha$ in our reward design is 0.0003 for MNIST permutations, 0.0002 for MNIST mix, and 0.001 for dataset CIFAR-100. The l1\_lambda is 0.00001, l2\_lambda is 0.0001, gl\_lambda is 0.001, regular\_lambda is 0.5, loss\_thr is 0.01, spl\_thr is 0.05 in DEN for MNIST permutations and MNIST mix. As for CIFAR-100, the hyperparameters in DEN is the same except regular\_lambda is 5.

\end{document}